# Energy Disaggregation via Deep Temporal Dictionary Learning

Mahdi Khodayar, *Student Member, IEEE*, Jianhui Wang, *Senior Member, IEEE,*
and Zhaoyu Wang, *Member, IEEE*

*Abstract*— This paper addresses the energy disaggregation problem, i.e. decomposing the electricity signal of a whole home to its operating devices. First, we cast the problem as a dictionary learning (DL) problem where the key electricity patterns representing consumption behaviors are extracted for each device and stored in a dictionary matrix. The electricity signal of each device is then modeled by a linear combination of such patterns with sparse coefficients that determine the contribution of each device in the total electricity. Although popular, the classic DL approach is prone to high error in real-world applications including energy disaggregation, as it merely finds linear dictionaries. Moreover, this method lacks a recurrent structure; thus, it is unable to leverage the temporal structure of energy signals. Motivated by such shortcomings, we propose a novel optimization program where the dictionary and its sparse coefficients are optimized simultaneously with a deep neural model extracting powerful nonlinear features from the energy signals. A long short-term memory auto-encoder (LSTM-AE) is proposed with tunable time dependent states to capture the temporal behavior of energy signals for each device. We learn the dictionary in the space of temporal features captured by the LSTM-AE rather than the original space of the energy signals; hence, in contrast to the traditional DL, here, a nonlinear dictionary is learned using powerful temporal features extracted from our deep model. Real experiments on the publicly available Reference Energy Disaggregation Dataset (REDD) show significant improvement compared to the state-of-the-art methodologies in terms of the disaggregation accuracy and F-score metrics.

*Index Terms*— Energy Disaggregation, Dictionary Learning, Long Short-term Memory Auto-encoder, Deep Learning

## I. INTRODUCTION

ENERGY disaggregation also known as non-intrusive load monitoring, is the problem of decomposing the whole electricity consumption signal of a residential, commercial, or industrial building into the signals of its appliances. The disaggregation models can inform the service customers of their consumption patterns and recognize malfunctions in electricity appliances [1]. Furthermore, finding the detailed electricity consumption patterns of the customers helps energy suppliers to efficiently plan and operate power systems.

Motivated by such beneficial applications, the energy society has been recently interested in this problem. Energy disaggregation studies are categorized into two groups. The first class of approaches classify electricity events rather than decomposing the energy signals. [2] considers each device as a finite state machine and identifies sharp edges of the active/reactive power. The appliances are further clustered based on their electricity changes. The devices with low energy consumption are likely to be assigned to the same cluster; hence, degrading the accuracy of this method. Later works including [3] leveraged transient and harmonic information with very high sampling rates; however, such data requires costly hardware and monitoring devices. Also, in this line of research, several load classification methods using different factors such as load control [4] or power usage [5] have been presented.

The second class of algorithms decompose the total electricity signal into its component devices [1, 6]. While the unsupervised models [7] do not require individual appliances' data for the disaggregation task, the supervised approaches make use of such data collected from the target building. [8] proposes a supervised discriminative sparse coding model based on structured prediction, to maximize disaggregation performance. The whole energy signal of each device is modeled as a sparse linear combination of an unknown dictionary; hence, this model requires lots of data to be trained. [9] introduces the factorial hidden Markov model (FHMM) using block Gibbs sampling for energy signal decomposition. [10] further enhances FHMM by an approximate algorithm based on convex programming in order to address the unmodeled devices. Although FHMM is flexible to be applied in both supervised/unsupervised settings, the EM training procedure of this model highly depends on initialization; hence, degrading the accuracy.

In this paper, the deep temporal dictionary learning (DTDL) is presented as a novel supervised algorithm for the problem of energy disaggregation. Given an aggregate signal of a whole building and a set of electricity signals for each device, we automatically capture powerful temporal consumption patterns for each device. To learn such patterns, a long short-term memory auto-encoder (LSTM-AE) is designed to capture the nonlinear manifold of energy signals for each device inside small time windows. Using a recurrent structure, we are able to extract useful temporal features for the underlying signals. Furthermore, an optimization program is proposed to tune the LSTM-AE's parameters while learning a dictionary of energy features that best represent the temporal features obtained by

Mahdi Khodayar and Jianhui Wang are with the Department of Electrical Engineering, Southern Methodist University, TX, USA (email: MahdiK@smu.edu and Jianhui@smu.edu).
Zhaoyu Wang is with the Department of Electrical and Computer Engineering, Iowa State University, IA, USA. (wzy@iastate.edu).

our auto-encoder. Learning the deep temporal features as well as the signal dictionary simultaneously, our model is able to capture a nonlinear dictionary that can leverage time dependent features of a deep model. Computing the dictionary, we disaggregate the building's consumption signal using our sparse coding model; hence, finding the contribution of each device to the aggregate electricity.

The contributions of the presented work are the following:
1) Previous dictionary learning works merely optimize for a linear dictionary, which makes them error-prone for real-world applications where most inputs/features have nonlinear characteristics and relationships. In this study, a nonlinear dictionary is computed for the electricity signals; hence, improving the accuracy of energy disaggregation solutions.
2) In contrast to previous sparse modeling works, DTDL finds the optimal dictionary and sparse codes in the space of a deep model's latent features rather than the original space of the energy signals; hence, our algorithm is able to leverage powerful deep learning features extracted from the electricity signals of various devices.
3) This is the first work that captures the temporal manifold of energy signals for each device; hence, learning power unsupervised features from energy signals while employing such features in the disaggregation task. Using the recurrent structure of our presented LSTM-AE, our approach is able to model the temporal structures of electricity patterns.

The rest of the paper is organized as the following: Section II defines the energy disaggregation problem. In Section III, the classic DL approach is explained; furthermore, the proposed DTDL model using LSTM-AE for temporal feature extraction is discussed. Section IV presents our novel optimization for nonlinear deep dictionary learning. Section V shows real experiments on a publicly available electricity dataset. Finally, Section VI presents the conclusion of this research.

## II. PROBLEM FORMULATION

Let us assume there are $L$ electric devices in a building and each device $i$ consumes an energy signal $x_i(t)$ at each time $1 \leq t \leq T$. The aggregate consumption signal observed (recorded) by the smart meter is computed by:

$$\bar{x}(t) = \sum_{i=1}^{L} x_i(t) \quad (1)$$

where $\bar{x}(t)$ is the total power consumed at time $t$. Observing the aggregated signals $\{\bar{x}(t)\}_{t=1}^{T}$, the goal is to recover the consumption signal of the individual appliances $1 \leq i \leq L$, i.e. the estimation of $\{x_i(t)\}_{t=1}^{T}$ for each valid $i$ and $t$.

Let us consider an energy consumption dataset $C$ corresponding to a building, that contains the energy signals of different devices through time (from $t = 1$ up to $t = T$). We break the consumption signals into windows of length $\omega \ll T$ for all devices and denote the consumption electricity of each device $i$ in the time interval $[(k-1)\omega + 1, k\omega]$ by $y_i(k)$, called an *energy snippet*, for all $k = 1, 2, ..., K = \frac{T}{\omega}$. The corresponding aggregate signal is denoted by $\bar{y}(k)$. As a result, $C$ is defined by $C = \{C_1, C_2, ..., C_K\}$ in which each data sample $C_k$ is a tensor of the form $\langle y_1(k), y_2(k), ..., y_L(k), \bar{y}(k) \rangle$. The goal is to build a dictionary matrix $D \in \mathbb{R}^{\omega \times N}$ such that a solution of the following problem reveals the disaggregation of $\bar{y}(k)$:

$$\bar{y}(k) = D\,a(k) = \sum_{j=1}^{N} a_j(k) D_{.,j} \quad (2)$$
$$D = [D_1\,D_2\,...\,D_L] \in \mathbb{R}^{\omega \times N} \quad D_i \in \mathbb{R}^{\omega \times N_i}$$

Here, $D$ is a dictionary matrix of size $\mathbb{R}^{\omega \times N}$. Each column $j$ of the dictionary, i.e. $D_{.,j}$, is a representative signal (also called an *atom*) for the device consumption signals $y_i(k)\ i \in [1, L]$, $k \in [1, K]$; that is, every signal $y_i(k)$ can be written as a linear combination of several columns (atoms) in $D$; $a(k)$ is a sparse coefficient vector that determines the coefficients of such a linear combination. Each element $a_j(k)$ decides the contribution of each column $D_{.,j}$ in the total consumption signal $\bar{y}(k)$. As shown in (1), we decompose $D$ into $L$ sub-dictionaries $D_i \in \mathbb{R}^{\omega \times N_i}$, each corresponding to a device; hence, each signal $y_i(k)\ k \in [1, K]$ can be written as a linear combination of columns of the sub-dictionary $D_i$, while $N_i$ is the number of these columns (atoms) defined for each device $i$. Therefore, the aggregate signal $\bar{y}(k)$ is a linear combination of the columns (atoms) in $D_i$ each associated with a sparse coefficient vector $a^i(k)$ written by:

$$\bar{y}(k) = D\,a(k)$$
$$D = [D_1\,D_2\,...\,D_L] \in \mathbb{R}^{\omega \times N} \quad (3)$$
$$a(k) = [a^1(k)\,a^2(k)\,...\,a^L(k)] \in \mathbb{R}^{N} \quad a^i(k) \in \mathbb{R}^{N_i}$$

Since each device has multiple consumption patterns corresponding to different operation modes, the objective is to extract useful consumption signatures (temporal patterns) through time to build sub-dictionaries $D_i$ for each device $i$, as a matrix whose columns (atoms) can best represent the energy snippets $y_i(k)\ k \in [1, K]$. Moreover, we need to find the optimal sparse coefficients $a^i(k)$ for all devices in order to reveal the contribution of each device in the total consumption signal $\bar{y}(k)$. Notice that the devices with a non-zero $a^i(k)$ are detected to be operating/on at the time interval $k$, while others are considered to be off.

## III. DICTIONARY LEARNING FOR ENERGY DISAGGREGATION

In this section, first, we discuss a classic dictionary learning (DL) method to solve (3) to find the optimal $D$ and $a(k)$, then, the drawbacks of this approach are explained and a novel nonlinear deep dictionary learning methodology is proposed for the energy disaggregation problem.

### A. Classic Dictionary Learning

One can find the optimal sparse coefficients $a^*(k)$ for each $k$ by solving the sparse coding problem with $l_1$ regularization as formulated by:

$$a^*(k) = \underset{a(k)}{\operatorname{argmin}} ||\bar{y}(k) - Da(k)||_2^2 + \lambda_1 ||a(k)||_1 \quad (4a)$$
$$\text{s.t.} \quad 1^T a^i(k) \leq 1 \quad a^i(k) \in \{0,1\}^{N_i} \quad (4b)$$

Here, $||\bar{y}(k) - Da(k)||_2^2$ is the reconstruction error that computes the distance between $Da(k)$ with the true total energy $\bar{y}(k)$, while $||a(k)||_1$ is the sparsity error that ensures the sparsity of $a(k)$. A sparsity coefficient $\lambda_1$ provides a trade-off between the reconstruction accuracy and sparsity of the solution $a^*(k)$. The condition in (4b) makes sure that for each device $i$, only one column (atom) is found as the signature of that device in the aggregate signal $\bar{y}(k)$; that is, for each device, only one column of $D_i$ is selected to have non-zero coefficient inside $a^*(k)$; thus, the device $i$ with a non-zero element in $a^i(k)$ is operating and its contribution to the total consumption $\bar{y}(k)$ is determined by:
$$(D_i)_{.,j} (a^i(k))_j \quad s.t. \quad (a^i(k))_j \neq 0 \tag{5}$$
where $(D_i)_{.,j}$ is the $j$-th column of $D_i$ while $(a^i(k))_j$ is the $j$-th entry of $a^i(k)$.

Furthermore, to find the optimal dictionary $D$ with respect to the dataset $C$ with $K$ data samples, one can minimize the following empirical cost function over the dictionary $D$ and sparse coefficient matrix $A = [a(1)\ a(2)\ ...\ a(K)] \in \mathbb{R}^{N \times K}$:

$$\min_{D,A} \frac{1}{K} \sum_{k=1}^{K} (\|\bar{y}(k) - Da(k)\|_2^2 + \lambda_1 \|a(k)\|_1) \tag{6a}$$
$$s.t.\ \|D_{.,j}\|_2^2 \leq 1 \quad j = 1,2,...,N \tag{6b}$$
$$1^T a^i(k) \leq 1\ a^i(k) \in \{0,1\}^{N_i}\ i = 1,2,...,L\ k = 1,2,...,K \tag{6c}$$

here, the constraint in (6b) prevents the dictionary from being arbitrarily large, since it can cause very small coefficient values in $A$, which makes the solution less informative.

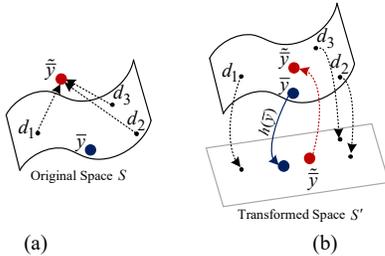

Fig. 1. (a) Classic DL: estimating the true consumption signal $\bar{y}(k)$ by dictionary atoms $d_1$, $d_2$ and $d_3$ inside the original nonlinear space $S$. (b) DTDL: Transformation of $S$ by a mapping function $h$ to learn dictionary atoms inside the transformed space $S'$. The mapping provides a better estimation $\bar{\bar{y}}(k)$ for $\bar{y}(k)$ when mapped back to $S$.

### B. Deep Temporal Dictionary Learning: A Novel Paradigm Towards Dictionary Learning for Energy Disaggregation

#### 1) Challenges and Contributions

The classic optimization in (4a)-(6c) has two major drawbacks that motivate the need for a more powerful framework for the problem of energy disaggregation:
1- Linearity of solution: the classic dictionary learning learns a linear $D$. As shown in Fig. 1, the minimization in the classic DL model finds an approximation $\bar{\bar{y}}(k)$ for the true value $\bar{y}(k)$ by finding dictionary atoms $D_{.,j}$ that are inside the original space of $y_i(k)\ i \in [1,L]\ k \in [1,K]$; however, if such a space is nonlinear (as in the case of most real-world applications including energy disaggregation), the estimation value $\bar{\bar{y}}(k)$ might not be in the original space $S$, making the estimation $Da(k)$ useless for modeling the true $\bar{y}(k)$. This motivates us to devise a novel non-linear dictionary learning based on deep learning to provide a nonlinear mapping from $S$ to an appropriate transformed space $S'$ in which $\bar{\bar{y}}(k)$ can be well written as a combination of atoms of $D$ as both $\bar{y}(k)$ and the columns $D_{.,j}$ lie on the same space $S'$. Learning such a nonlinear mapping, i.e. learning the transformed space $S'$, is a crucial challenge solved in this Section. 2- Classic DL algorithms lack a recurrent structure to model the temporal behavior of the underlying energy consumption dataset; thus, the need for a recurrent optimization model that can capture useful temporal patterns from the underlying data, i.e. signals $y_i(k)$ inside the dataset $C$, is raised. As the energy consumption signals are all time series, we propose a recurrent optimization model to address this issue.

#### 2) Proposed Deep Recurrent Model

To tackle the presented challenges in Section III-B-1, we propose a novel deep learning based optimization for energy disaggregation. Our method is a DL approach with a deep recurrent formulation to capture nonlinear temporal features that can help our model to better understand the behavior of the energy consumption temporal data, i.e. $y_i(k)$ signals. Our main idea is that in order to learn each sub-dictionary $D_i$ corresponding to each device $i$ and the sparse code matrix $A$, we learn $N_i$ number of optimal $\omega$−dimensional energy snippets $\hat{y}_i = \langle \hat{y}_i(1), \hat{y}_i(2), ..., \hat{y}_i(N_i) \rangle \in \mathbb{R}^{\omega \times N_i}$ for each device $i$, such that the elements of $\hat{y}_i$ best represent the energy snippets $y_i(k)$ for $k \in [1,K]$. In other words, for each device $i$, every $y_i(k)$ can be written as a linear combination of elements of $\hat{y}_i$; hence, one can conclude that the optimal sub-dictionary is found by $D_i = \hat{y}_i \in \mathbb{R}^{\omega \times N_i}$.

Assuming that the consumption snippets $y_i(k)\ k \in [1,K]$ lie on a nonlinear manifold $M$, in order to find $D_i = \hat{y}_i$, we learn a nonlinear transformation $F_{enc}: \mathbb{R}^\omega \to \mathbb{R}^d$ that encodes each energy snippet $y_i(k) \in \mathbb{R}^\omega$ by a $d$-dimensional feature vector $h(y_i(k)) \in \mathbb{R}^d$ that captures the fundamental nonlinear temporal features of the energy snippet $y_i(k)$. The feature $h(y_i(k))$ is further decoded by a nonlinear mapping $F_{dec}: \mathbb{R}^d \to \mathbb{R}^\omega$ that maps the learned feature $h(y_i(k))$ in the nonlinear (transformed) space to the observed $y_i(k)$ in the original space; hence, learning a powerful nonlinear mapping $h$ that is capable of reconstructing the original consumption signal. Such nonlinear mapping $h$ is implemented by $F_{enc}$ and mapped back to the original space of energy snippets by $F_{dec}$. While $h(y_i(k))$ is learned (i.e. $F_{enc}$ and $F_{dec}$ are found) for all $y_i(k)$ in the dataset $C$, our model learns $D_i = \hat{y}_i$ inside the transformed space corresponding to the nonlinear mapping $h$; hence learning the nonlinear dictionary $D_i = \hat{y}_i$ for the energy snippets $y_i(k)\ k \in [1,K]$ for each device $i$.

Since we are working with temporal data $y_i(k)$, we devise a long short-term memory auto-encoder (LSTM-AE) neural network using a deep learning based recurrent formulation. As shown in Fig. 2, our LSTM-AE is an LSTM network with $2\omega$ temporal states $S_i\ i = 1,2,...,2\omega$. The first $\omega$ states serve to model $F_{enc}$ that maps $y_i(k)$ to $h(y_i(k))$ in $\omega$ iterations. At each iteration $1 \leq l \leq \omega$, an element $y_i(k)_l \in \mathbb{R}$ is observed by the LSTM unit and the temporal state $S_{l-1}$ is updated to $S_l$ using the following recurrent formulation:

$$\begin{aligned}
x_l &= y_i(k)_l \\
a_l &= \tanh(W_a x_l + U_a h_{l-1} + b_a) \\
i_l &= \text{Sigm}(W_i x_l + U_i h_{l-1} + b_i) \\
f_l &= \text{Sigm}(W_f x_l + U_f h_{l-1} + b_f) \\
o_l &= \text{Sigm}(W_o x_l + U_o h_{l-1} + b_o) \\
S_l &= f_l \circ S_{l-1} + i_l \circ a_l \\
h_l &= o_l \circ \tanh(S_l)
\end{aligned} \tag{7}$$

where $x_l$ is the input vector, while $i_l$, $f_l$, and $o_l$ are the $m$-dimensional input gate, forget gate, and output gate decision variables at iteration $l$, respectively. $a_l$ is the input activation with bias $b_a$, and $h_l$ is the output vector of the LSTM at iteration $l$, which stores the temporal features obtained from the input sequence from the iterations 1 up to $l$. The parameters $b_i, b_f, b_o,$ and $b_S$ are the $m$- dimensional bias vectors while $W_i$, $W_f$, $W_o$, and $W_S$ are weight parameters inside $\mathbb{R}^m$; Moreover, $U_i$, $U_f$, $U_o$, and $U_S$ are the weight matrices in $\mathbb{R}^{m \times m}$. All bias and weight parameters are tunable parameters that are learned to find the optimal state $S_l$ as well as the optimal temporal feature $h_l$ at each iteration $l$. When $l = \omega$, the whole $y_i(k)$ signal has been observed and $h_l = h_\omega = h(y_i(k))$ is obtained by the LSTM as the temporal features of the whole consumption signal $y_i(k)$; thus, the first $\omega$ iterations of the LSTM implement $F_{enc}$ mapping each energy snippet $y_i(k)$ to the corresponding temporal feature.

As shown in Fig. 2, the iterations $\omega + 1 \leq l \leq 2\omega$ reconstruct $y_i(k)$; At each $l$, an output feature $h_l = y_i(k)_{l-\omega}$ is generated by the LSTM in order to model $F_{dec}$ that maps the resulting temporal features of $F_{enc}$, i.e. $h(y_i(k))$, to the original consumption snippet $y_i(k)$. This leads to learning the nonlinear temporal manifold of the energy snippets $y_i(k)$ $i = 1,2, \ldots, L$ $k = 1,2, \ldots, K$ as the LSTM learns powerful temporal features $h_\omega = h(y_i(k))$ in its $l = \omega$ iteration, that are so powerful that can reconstruct the original energy snippets $y_i(k)$.

*3) Proposed Optimization of Dictionary Learning for Energy Disaggregation*

We learn the sub-dictionaries $D_i$ and the sparse coefficient matrix $A$ in the space of $h_\omega = h(y_i(k))$; that is, when $y_i(k)$ is mapped in the $\omega$-th iteration of the LSTM to the feature vector $h_\omega$, the columns (atoms) of the corresponding sub-dictionary $D_i$ are learned to represent $h_\omega$ so that the linear combinations of the atoms (columns) of each $D_i$ would be able to yield the feature vectors $h_\omega = h(y_i(k))$ for all $k = 1,2, \ldots, K$. Such linear combinations are determined by the sparse code matrix $A$.

We define the following optimization program for the problem of learning the optimal $D$ and $A$, while finding the optimal mappings $F_{enc}$ and $F_{dec}$:

$$\min_{F_{enc},F_{dec},\{D_i\}_{i=1}^L,A} J = J_1 + \lambda_2 J_2 + \lambda_3 J_3 + \lambda_4 J_4$$

$$J_1 = \frac{1}{L}\sum_{i=1}^L \frac{1}{K}\sum_{k=1}^K \left(\|F_{enc}(y_i(k)) - D_i a^i(k)\|_2^2 + \lambda_1 \|a^i(k)\|_1\right)$$

$$J_2 = \sum_{j=1, j \neq i}^L \|D_i^T D_j\|_F^2$$

$$J_3 = \frac{1}{L}\sum_{i=1}^L \frac{1}{K}\sum_{k=1}^K \left(\|F_{dec}(F_{enc}(y_i(k))) - y_i(k)\|_2^2\right) \quad (8)$$

$$J_4 = \begin{pmatrix} \|W_f\|_F^2 + \|W_i\|_F^2 + \|W_o\|_F^2 + \|U_f\|_F^2 + \|U_i\|_F^2 + \|U_o\|_F^2 \\ + \|b_f\|_2^2 + \|b_i\|_2^2 + \|b_o\|_2^2 \end{pmatrix}$$

$$\text{s.t.} \quad \frac{1}{K}\sum_{k=1}^K |1^T a^i(k) - 1^T a^i(k+1)| = 0 \quad \forall i$$

$$\|(D_i)_{.,j}\|_2^2 \leq 1 \quad i = 1,2, \ldots, L \quad j = 1,2, \ldots, N_i$$

Here, $J_1$ is the dictionary learning cost function to compute the difference between $h_{l=\omega} = h(y_i(k)) = F_{enc}(y_i(k))$ and the linear combination of the atoms in the sub-dictionary $D_i$ computed by $D_i a^i(k)$. Such difference is considered for all devices $i$ and all time intervals $k$ in the dataset. $\lambda_1 \|a^i(k)\|_1$ ensures sparsity for the solution of $A$. $J_2$ is our cross-dictionary incoherence term; minimizing this error promotes incoherence between two sub-dictionaries $D_i$ and $D_{j \neq i}$; that is, this error term is in favor of having distinct sub-dictionary atoms for different devices. $J_3$ is the LSTM-AE's reconstruction error term, which is the distance between the output of LSTM-AE generated at iterations $\omega + 1 \leq l \leq 2\omega$, i.e. $F_{dec}(F_{enc}(y_i(k)))$, and the desired output $y_i(k)$ for all devices $i$ and all time intervals $k$. $J_4$ is the regularization error defined to control the magnitude of the LSTM-AE's parameters. Large parameters might lead to the overfitting problem; $J_4$ is defined over LSTM parameters of (7) in order to avoid such problem. $J_5$ is an error term to satisfy the temporal smoothness prior. Notice that, for any device $i$, the term $|1^T a^i(k) - 1^T a^i(k+1)|$ is zero except at intervals when it turns on/off. Given the fact that such switching happens in very small periods of time compared to the whole time period, we minimize $\frac{1}{K}\sum_{k=1}^K |1^T a^i(k) - 1^T a^i(k+1)|$ for all devices. Finally, the constraint $\|(D_i)_{.,j}\|_2^2 \leq 1$ is assumed to avoid each sub-dictionary from obtaining arbitrary large entries as it would cause very small coefficient matrix $A$, making the solution trivial.

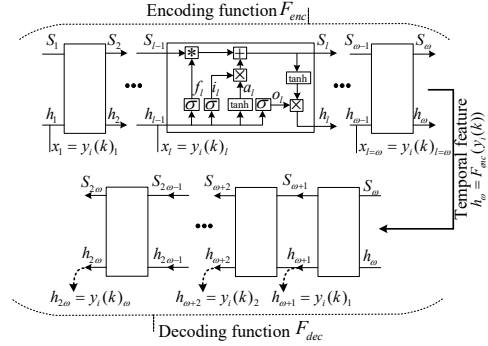

Fig. 2. Structure of the proposed long short-term memory auto-encoder

## IV. OPTIMIZATION SOLUTION FOR ENERGY DISAGGREGATION

The optimization program in (8) is not jointly convex with respect to $F_{enc}$, $F_{dec}$, $\{D_i\}_{i=1}^L$, and $A$. As a result, we present an iterative algorithm that alternates between optimizing the functions $F_{enc}$ and $F_{dec}$, as well as the variables $\{D_i\}_{i=1}^L$ and $A$. Our proposed algorithm addresses three sub-problems alternately as explained in the following.

*A. Optimize the LSTM-AE mappings $F_{enc}$ and $F_{dec}$:*

Having a fixed $D$ and $A$, in order to optimize $F_{enc}$ and $F_{dec}$ in our objective (8), we need to solve the following optimization:

$$\min_{F_{enc},F_{dec},} \tilde{J} = \left\{\frac{1}{L}\sum_{i=1}^L \frac{1}{K}\sum_{k=1}^K \left[\left(\|F_{enc}(y_i(k)) - D_i a^i(k)\|_2^2\right) + \left(\|F_{dec}(F_{enc}(y_i(k))) - y_i(k)\|_2^2\right)\right]\right\}$$

$$+ \lambda_4 \begin{pmatrix} \|W_f\|_F^2 + \|W_i\|_F^2 + \|W_o\|_F^2 + \|U_f\|_F^2 + \|U_i\|_F^2 + \|U_o\|_F^2 \\ + \|b_f\|_2^2 + \|b_i\|_2^2 + \|b_o\|_2^2 \end{pmatrix} \quad (9)$$

As explained in Section 3-B-2, the term $F_{enc}(y_i(k))$ in (9) is the output of the $\omega$-th iteration of the LSTM-AE denoted by $h_{l=\omega}$; hence, we need to train our LSTM unit to output $h_{l=\omega} = D_i a^i(k)$ at this iteration. Moreover, $F_{dec}(F_{enc}(y_i(k)))$ is the output of the LSTM-AE in iterations $\omega + 1$ through $2\omega$, i.e. $F_{dec}(F_{enc}(y_i(k))) = [h_{\omega+1} \; h_{\omega+2} \; \ldots \; h_{2\omega}]$; Therefore, we need to train LSTM-AE to satisfy $[h_{\omega+1} \; h_{\omega+2} \; \ldots \; h_{2\omega}] = y_i(k)$ in (9). Let us define the following notations for our LSTM parameters at

iteration $l$:

$$gates_l = \begin{bmatrix} a_l \\ i_l \\ f_l \\ o_l \end{bmatrix}, \quad W = \begin{bmatrix} W_a \\ W_i \\ W_f \\ W_o \end{bmatrix}, \quad U = \begin{bmatrix} U_a \\ U_i \\ U_f \\ U_o \end{bmatrix}, \quad b = \begin{bmatrix} b_a \\ b_i \\ b_f \\ b_o \end{bmatrix} \quad (10)$$

To minimize $\tilde{J}$ in (9) after observing each $y_i(k)$, we compute the gradient of $\tilde{J}$ with respect to the LSTM's output $h_l$ using:

$$\Delta_l = \frac{\partial \tilde{J}}{\partial h_l}(y_i(k)) = \begin{cases} 2[F_{enc}(y_i(k)) - D_i a^i(k)] & l = \omega \\ 2[[h_{\omega+1}\ h_{\omega+2}\ \ldots\ h_{2\omega}] - y_i(k)] & l \geq \omega + 1 \end{cases} \quad (11)$$

hence, for each LSTM iteration $l = 1,2,\ldots,\omega$, we compute the partial derivatives of $\tilde{J}$ with respect to various LSTM's gates by:

$$\begin{aligned}
\delta h_l &= \Delta_l + \Delta h_l \\
\delta S_l &= \delta h_l \circ o_l \circ (1 - \tanh^2(S_l)) + \delta S_{l+1} \circ f_{l+1} \\
\delta a_l &= \delta S_l \circ i_l \circ (1 - a_l^2) \\
\delta i_l &= \delta S_l \circ a_l \circ i_l \circ (1 - i_l) \\
\delta f_l &= \delta S_l \circ S_{l-1} \circ f_l \circ (1 - f_l) \\
\delta o_l &= \delta h_l \circ \tanh(S_l) \circ o_l \circ (1 - o_l) \\
\delta x_l &= W^T \cdot \delta gates_l \\
\Delta h_{l-1} &= U^T \cdot \delta gates_l
\end{aligned} \quad (12)$$

Having (9)-(12), we compute the partial derivatives of $\tilde{J}$ with respect to LSTM's parameters $W, U,$ and $b$:

$$\begin{aligned}
\delta W &= \sum_{l=0}^{2\omega} \delta gates_l \otimes x_l + \lambda_4 W \\
\delta U &= \sum_{l=0}^{2\omega-1} \delta gates_{l+1} \otimes h_l + \lambda_4 U \\
\delta b &= \sum_{l=0}^{2\omega} \delta gates_{l+1} + \lambda_4 b
\end{aligned} \quad (13)$$

Considering (12), we update our LSTM-AE model (which is an implementation of $F_{enc}$ and $F_{dec}$) using the following update rule based on the gradient descent method:

$$\begin{aligned}
W^{new} &\leftarrow W - \eta \delta W \\
U^{new} &\leftarrow U - \eta \delta U \\
b^{new} &\leftarrow b - \eta \delta b
\end{aligned} \quad (14)$$

Here, $W^{new}$, $U^{new}$, and $b^{new}$ are the updated parameters $W, U,$ and $b$ using the gradient descent update rule (13) after observing each $y_i(k)$ $i = 1,2,\ldots,L$ $k = 1,2,\ldots,K$ in $2\omega$ iterations. $\eta$ is the learning rate that determines how strong each update can be.

*B. Optimize the Dictionary D:*

Given the fixed mappings $F_{enc}$ and $F_{dec}$, as well as some fixed sparse code matrix $A$, our optimization (8) will have the following form by which we seek to optimize $D$:

$$\min_{D=[D_1\ D_2\ \ldots\ D_L]} \bar{J} = \frac{1}{L}\sum_{i=1}^{L} \frac{1}{K}\sum_{k=1}^{K} \left(\left\|F_{enc}(y_i(k)) - D_i a^i(k)\right\|_2^2\right) + \lambda_2 \sum_{j=1, j\neq i}^{L} \|D_i^T D_j\|_F^2 \quad (15)$$

$$s.t.\ \|(D_i)_{.,j}\|_2^2 \leq 1 \quad i = 1,2,\ldots,L \quad j = 1,2,\ldots,N_i$$

Here, the cross sub-dictionary incoherence error term $\lambda_2 \sum_{j=1,j\neq i}^{L} \|D_i^T D_j\|_F^2$ tries to enforce the resulting sub-dictionaries of different devices $i \neq j$ to have distinct dictionary atoms. In order to investigate the effect of such error term on the accuracy of our solution, we assume two different settings to solve (15) by setting the coefficient $\lambda_2$ to a zero or non-zero value:

*1) No sub-dictionary incoherence error ($\lambda_2 = 0$) in (15):*

In this setting, there is no incoherence error; hence, each two devices might contain similar atoms in their corresponding sub-dictionaries. This changes the optimization of (15) to a least squares problem with quadratic constraints; thus, we solve it using Lagrangian multipliers. First, let us define the Lagrangian in the following form using Lagrangian multipliers $\phi = [\phi_{i,j} \geq 0]\ i = 1,2,\ldots,L\ j = 1,2,\ldots,N_i$:

$$\mathcal{L}(D,\phi) = \frac{1}{L}\sum_{i=1}^{L} \frac{1}{K}\sum_{k=1}^{K} \left(\left\|F_{enc}(y_i(k)) - D_i a^i(k)\right\|_2^2\right) + \sum_{i=1}^{L}\sum_{j=1}^{N_i} \phi_{i,j}\left(\|(D_i)_{.,j}\|_2^2 - 1\right) \quad (16)$$

Considering the Lagrangian multipliers as $\tilde{\phi} = [\tilde{\phi}_j \geq 0]_{j=1}^{N}$, one can rewrite (16) using:

$$\mathcal{L}(D,\phi) = \frac{1}{L}\sum_{i=1}^{L} \frac{1}{K}\sum_{k=1}^{K} \left(\left\|F_{enc}(y_i(k)) - D_i a^i(k)\right\|_2^2\right) + \sum_{j=1}^{N} \tilde{\phi}_j \left(\|D_{.,j}\|_2^2 - 1\right) \quad (17)$$

Having $\frac{\partial \mathcal{L}(D,\phi)}{\partial D} = 0$, we find the following analytical solution:

$$D = F_{enc}^{Total} A^T (AA^T + \Sigma)^{-1}$$

$$F_{enc}^{Total} = \langle \begin{matrix} F_{enc}(y_{i=1}(1)) \ldots F_{enc}(y_{i=1}(K)) \\ \ldots \\ F_{enc}(y_{i=L}(1)) \ldots F_{enc}(y_{i=L}(K)) \end{matrix} \rangle \in \mathbb{R}^{d \times (L*K)} \quad (18)$$

where $F_{enc}^{Total} \in \mathbb{R}^{d \times (L*K)}$ is a vector of all $F_{enc}(y_{i=l}(k))$ for all $i = 1,2,\ldots,L$ and $k = 1,2,\ldots,K$. Notice that $d = \dim(h_{l=\omega})$ is the dimension of the temporal feature vector $h_{l=\omega} = F_{enc}(y_i(k))$; also, $\Sigma = (L * K)diag(\phi) \in \mathbb{R}^{N \times N}$. Therefore, the corresponding Lagrangian dual function is written as:

$$\mathcal{L}_{dual}(\phi) = \min_D \mathcal{L}(D,\phi)$$
$$= \frac{1}{L}\sum_{i=1}^{L}\frac{1}{K}\sum_{k=1}^{K} \|F_{enc}(y_i(k)) - F_{enc}^{Total} A^T (AA^T + \Sigma)^{-1} a^i(k)\|_2^2 \quad (19)$$
$$+ \sum_{j=1}^{N} \tilde{\phi}_j (\|F_{enc}^{Total} A^T (AA^T + \Sigma)^{-1} u_i\|_2^2 - 1)$$

with the $i$-th unit vector denoted by $u_i \in \mathbb{R}^N$. Leveraging the gradient descent method, we maximize the dual Lagrangian $\mathcal{L}_{dual}(\phi)$ in (19) with respect to $\tilde{\phi} = [\tilde{\phi}_j \geq 0]_{j=1}^{N}$. The gradient of the dual for any $\tilde{\phi}_j$ is computed by:

$$\frac{\partial \mathcal{L}_{dual}(\phi)}{\partial \tilde{\phi}_j} = \|F_{enc}^{Total} A^T (AA^T + \Sigma)^{-1} u_i\|_2^2 - 1 \quad (20)$$

when the optimal $\tilde{\phi}$ is computed using gradient descent, the optimal dictionary $D$ is computed by (18) using the optimal $\Sigma = (L * K)diag(\phi)$.

*2) Consider sub-dictionary incoherence error ($\lambda_2 \neq 0$) in (15):*

When the sub-dictionary incoherence error term is considered in our optimization, i.e. $\lambda_2 \neq 0$, we need to optimize (15). Applying gradient descent, we minimize $\bar{J}$ in (15) with respect to each sub-dictionary for each training data $y_i(k)$ using the following gradient value for each sub-dictionary $D_i$:

$$\frac{\partial \bar{J}}{\partial D_i}(y_i(k)) = D_i a^i(k) \left(a^i(k)\right)^T - F_{enc,i} \left(a^i(k)\right)^T + \lambda_2 \sum_{j=1, j\neq i}^{L} (D_i^T D_j) D_i \quad (21)$$

$$F_{enc,i} = \langle F_{enc}(y_i(1)) \ldots F_{enc}(y_i(K)) \rangle \in \mathbb{R}^{d \times K}$$

Using gradient descent, one can minimize the optimization error $\bar{J}$ with respect to every sub-dictionary $D_i$; hence, optimizing the whole dictionary $D$.

*C. Optimize the Sparse Code Matrix A:*

When $F_{enc}$, $F_{dec}$, and $D$ are fixed, one can optimize the coefficient matrix $A$ while observing each signal $y_i(k)$. Let us write our main optimization in (8) in the following form where the main objective $J$ in (8) is optimized with respect to each $a^i(k)$ in $A$:

$$\min_{a^i(k)} \bar{J} = \left\|F_{enc}(y_i(k)) - D_i a^i(k)\right\|_2^2 + \lambda_1 \|a^i(k)\|_1 \quad (22)$$

$$s.t.\ |1^T a^i(k) - 1^T a^i(k+1)| = 0 \quad \forall i,k$$

Here, to satisfy the constraint $|1^T a^i(k) - 1^T a^i(k+1)| = 0$ for all $i$ and $k$, let us rewrite this condition using a square matrix $G$:

$$A\,G = 0 \quad A \in \mathbb{R}^{N\times(K*L)} \quad G \in \mathbb{R}^{(K*L)\times(K*L)}$$
$$A = \begin{pmatrix} a^{i=1}(1)\ a^{i=2}(1) \ldots a^{i=L}(1) \ldots a^{i=1}(K)\ a^{i=2}(K) \ldots a^{i=L}(K) \end{pmatrix}$$
$$= (A_{.,1}\ A_{.,2} \ldots A_{.,j} \ldots A_{.,(K*L)}) \quad j = (k-1)L + i \quad (23)$$
$$G_{i,j} = \begin{cases} 1 & i = j \\ -1 & i = j+L \\ 0 & \text{otherwise} \end{cases}$$

Having the constraint $A\,G = 0$, i.e. $G^T A^T = 0$, one can rewrite the optimization (22) to solve for each column $a^i(k) = A_{.,j}$ $j = (k-1)L + i$ for all energy snippets $y_i(k)$ by:

$$\min_{A_{.,j}=(k-1)L+i} \sum_{i=1}^{L} \sum_{k=1}^{K} \left\| F_{enc}(y_i(k)) - D A_{.,j} \right\|_2^2 + \lambda_1 \|A_{.,j}\|_1$$
$$\text{s.t.} \sum_{i'=1}^{K*L} \sum_{j'=1}^{K*L} G_{j',i'} A_{.,j} = 0 \quad (24)$$

Solving (24) by Proximal Jacobian ADMM [11], the optimal sparse coefficient matrix is computed. Notice that, the dimension of $G$ does not add much computational burden to our optimization program as a large portion of $G$'s entries are zero.

*D. Energy Disaggregation Algorithm Using the Proposed Optimization*

Algorithm 1 shows the structure of our energy disaggregation algorithm using the proposed dictionary learning model. We optimize $F_{enc}$, $F_{dec}$, $\{D_i\}_{i=1}^L$, and $A$ using an iterative algorithm alternating among the optimizations (9), (15) and (24). The optimizations are repeated until the average change in the dictionary entries is less than a small $\varepsilon > 0$.

In the test time, the optimal dictionary $D^*$ is used in (25) to obtain the optimal coefficients $a^*$ for some test aggregate energy signal $Y$ of a building. Having the optimal dictionary $D^*$, the optimal coefficients $a^*$ show the contribution of each device in the total electricity consumption $Y$. One can simply compute such contributions using (5).

---

**Algorithm 1** Deep Temporal Dictionary Learning for Energy Disaggregation

**Inputs:** consumption signals of all devices $y_i(k)$ $k \in [1, K]$ $i \in [1, L]$
**Outputs:** Optimal Dictionary $D = D^*$ and solution $a = a^*$ for the energy disaggregation problem $Y = Da$ where $Y$ is a test aggregate consumption signal that we decompose
1: **Repeat:**
2:   Optimize (9) to update $F_{enc}$, $F_{dec}$
3:   Optimize (15) to update the dictionary $D$
4:   Optimize (24) to update the sparse coefficients $A$
5: **Until** Convergence (changes in dictionary entries are less than $\varepsilon > 0$)
6: Test the model: Given the optimal dictionary $D^*$, compute optimal coefficient vector $a^*$ for an aggregate energy consumption signal $Y$:
$$a^* = \min_a \|F_{enc}(Y) - D^* a\|_2^2 + \lambda_1 \|a\|_1 \quad (25)$$

---

## V. SIMULATION RESULTS

*A. Dataset*

Our proposed energy disaggregation model, DTDL, is evaluated on the real-world REDD dataset [12], a large publicly available dataset for electricity disaggregation problems. The dataset contains power consumption signals of five houses with around 20 different appliances at each house. The electricity signals of each device as well as the total consumption are available for two weeks with a high frequency sampling rate of 15 kHz.

Knowing that the low frequency sampling leads to a more practical energy measurement that is less costly and more challenging for energy disaggregation, we train and evaluate our method on low frequency data. In this study, a sampling rate of 1Hz is applied to collect the energy signals. We train and validate DTDL using the data corresponding to the first week; 80% of these samples are used to train and the rest are used to validate the model to find the optimal hyperparameters. The samples of the second week are applied to test the model.

*B. Disaggregation Accuracy Metrics*

Let us assume the test aggregate consumption signal $Y$ contains $K$ time intervals (windows) of length $\omega$, each denoted by $Y(k)$ $k = 1,2,\ldots,K$. Signal $Y(k)$ is the summation of energy signals (energy snippets) $Y_i(k)$ $i = 1,2,\ldots,L$; that is, each device $1 \leq i \leq L$ consumes $Y_i(k)$ at the time interval $k$. Also, let us denote the estimation of $Y_i(k)$ by $\hat{Y}_i(k) = D_i\, a^i(k)$ obtained by our disaggregation method. The disaggregation accuracy is computed by:

$$acc = \left(1 - \frac{\sum_{k=1}^{K}\sum_{i=1}^{L} \left\|\hat{Y}_i(k) - Y_i(k)\right\|_1}{2 \sum_{k=1}^{K} \|Y(k)\|_1}\right) \times 100\% \quad (26)$$

Here, the 2 factor in the denominator is due to the fact that the absolute value leads to double counting errors.

In order to have a comprehensive comparison, we also compute precision, recall, and the F-score at device level. At each time period $k$, a binary "on/off" value indicates whether each device $i$ is operating ($a^i(k)$ is non-zero) or not ($a^i(k)$ is zero). Precision $P$ determines what portion of the estimated on/off decisions for a device truly belongs to that device, while recall $R$ measures what portion of the on/off value for one device is correctly estimated. F-score is the harmonic mean of $P$ and $R$ that combines these two metrics by:

$$F_{score} = \frac{2 \times P \times R}{P + R} \quad (27)$$

*C. Experimental Settings and Model Validation*

The learning rate $\eta$ of the LSTM-AE's update rule (14) is set to be 0.01 and the dictionary convergence threshold $\varepsilon$ in Algorithm 1 is set to be 0.05. LSTM-AE's hidden layer dimension $m$ and window length $\omega$ are determined by model validation; that is, we compute $acc$ on the validation dataset using different configurations of $m \in \{5,7,9,11,13\}$ and $\omega \in \{8,10,12,14,16\}$. The configuration with the highest $acc$ is chosen to test the model. For each device $i$, the number of dictionary atoms $N_i$ is set to be 20; hence, for a house with 20 appliances, a dictionary with $N = 20 \times 20 = 400$ columns is learned. Fig. 3 shows the validation $acc$ of DTDL averaged over all houses; As shown in this plot, the optimal configuration has $m = 7$ with a disaggregation accuracy of 83.56%. Increasing $m$ to larger values would grow the generalization capability (nonlinearity) of LSTM unit; however, it would also make the LSTM prone to overfitting; therefore, the moderate value of $m = 7$ is the optimal choice. It is also shown that the window size $\omega = 14$ leads to the highest validation accuracy. Notice that smaller windows would degrade the accuracy as the transients would be overemphasized when learning $D$; Moreover, larger time windows would lead to observing different dynamics/operation modes in just a single time window, hence decreasing the disaggregation accuracy.

In order to analyze the contribution of different error terms ($J_1, J_2, J_3$, and $J_4$) defined in the optimization (8) to the quality and accuracy of our energy disaggregation solution, we show the validation accuracy using different combinations of error coefficients $\lambda_2, \lambda_3$, and $\lambda_4$. Fig. 4 shows the validation $acc$ averaged over all houses for such configurations of our objective function $J$. In this plot, each error coefficient is chosen from $\{0, 0.2, 0.4, 0.6, 0.8, 1, 1.2, 1.4\}$. As shown in Fig. 4, the optimal configuration is $\langle \lambda_2, \lambda_3, \lambda_4 \rangle = \langle 0.4, 1.2, 0.6 \rangle$.

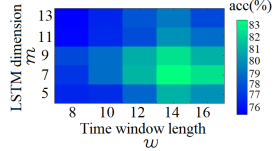

Fig. 3. Validation accuracy of DTDL with different configurations of LSTM dimension and window length

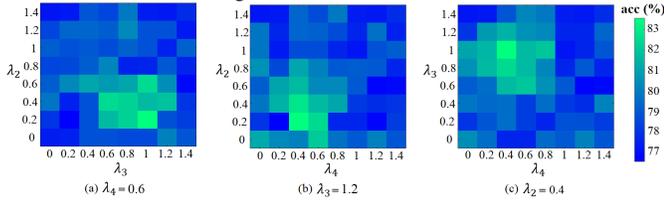

Fig. 4. Validation accuracy of DTDL with different configurations of error coefficients $\lambda_2, \lambda_3$, and $\lambda_4$. Contribution of various error terms to the accuracy of the disaggregation model is shown in terms of the accuracy matric in (26).

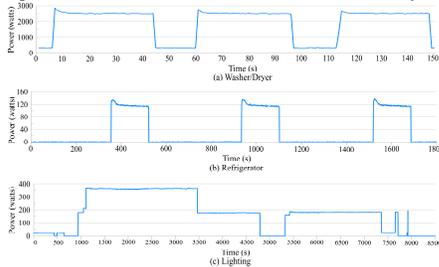

Fig. 5. The consumption pattern of wahser/dryer, refrigerator, and lighting device during their operation time for house 3 in the REDD dataset.

As discussed in the proposed optimization in Section III-B-3, $J_2$ is the cross dictionary incoherence error that enforces the devices to have dissimilar consumption patterns. The optimal coefficient $\lambda_2 = 0.4$ shows that this error term should have a relatively low (but more than zero) value, which means that, for some devices the assumption of having dissimilar energy snippets is true (e.g. refrigerator and lighting devices) while for other devices, the energy snippets might have similar behaviors. For instance, as shown in Fig. 5, the devices with rotary components (i.e. motors) such as refrigerator and washer/dryer have similar consumption patterns that are different from lighting appliances.

$J_3$ is the reconstruction error term that makes sure that the LSTM-AE has learned good temporal features that are strong enough to make (reconstruct) the original consumption signals. The optimal coefficient of this error term is $J_3 = 1.2$ which is relatively high. This shows that learning powerful temporal features help the model to find more accurate disaggregation solutions. Therefore, the high value of $J_3$ justifies our proposed approach for learning the transformed space $S'$ using the proposed LSTM-based model. Moreover, the regularization coefficient $J_4$ has a moderate value of 0.6 that shows the magnitude of the parameters of the LSTM-AE that need to be constrained in order to avoid overfitting.

### D. Numerical Results

We compare the proposed model, DTDL, with several energy disaggregation benchmarks including Simple Mean Prediction (SMP) [6], Factorial Hidden Markov Model (FHMM) [12], Approximate MAP Inference (AMAPI) [13], Hierarchical FHMM (HieFHMM) [14] and Powerlet-based Energy Disaggregation (PED) [6]. Moreover, the Classic Dictionary Learning (CDL) model discussed in Section III-A is considered as a baseline to better show the merit of deep learning in the area of sparse coding and dictionary learning.

Table I shows the energy disaggregation accuracy (26) of all benchmarks in the REDD dataset. As shown Table I, the dictionary learning models PED and DTDL have generally better performance compared to other methodologies. FHMM and its variants, AMAPI and HieFHMM, are outperformed by PED and DTDL since HMM-based models are limited by their first-order Markov property which makes them unable to capture high order correlation among various devices' consumption patterns. DTDL obtains the highest accuracy with 20.2%, 19.8% and 16.9% improvement over FHMM, AMAPI, and HieFHMM, respectively. The superiority of DTDL over the benchmarks is due to learning useful nonlinear patterns from electricity signals while incorporating the learned deep features in its dictionary learning process. Moreover, the recurrent structure of DTDL makes it a more powerful temporal pattern recognition model for the time dependent energy data.

TABLE I
DISAGGREGATION ACCURACY OF VARIOUS BENCHMARKS

| Methods | House | | | | | |
|---|---|---|---|---|---|---|
| | 1 | 2 | 3 | 4 | 5 | Average |
| SMP | 41.4 | 39.0 | 46.7 | 52.7 | 33.7 | 42.7 |
| FHMM | 71.5 | 59.6 | 59.6 | 69.0 | 62.9 | 64.5 |
| AMAPI | 73.2 | 61.4 | 62.3 | 60.1 | 66.3 | 64.7 |
| HieFHMM | 75.6 | 73.4 | 60.5 | 51.2 | 71.0 | 66.3 |
| CDL | 74.0 | 61.8 | 60.0 | 59.2 | 78.4 | 66.7 |
| PED | 81.6 | 79.0 | 61.8 | 58.5 | 79.1 | 72.0 |
| DTDL | **83.1** | **84.5** | **62.8** | **73.2** | **83.9** | **77.5** |

Table II shows the precision, recall, and F-score of all benchmarks. Let us compare the dictionary learning-based models: CDL, PED, and DTDL. On average, DTDL has 13.61% and 7.10% better F-score compared to CDL and PED, respectively. As explained in Section III-A, CDL learns a linear dictionary using the consumption signals of all devices. However, PED runs a dissimilarity-based subset selection model on the temporal windows of each device to find the windows that are most representative (windows that can best represent the whole set of windows). The representative windows of each device are used as the columns of the sub-dictionary corresponding to that device. In contrast to both CDL and PED, our DTDL approach learns a nonlinear dictionary that takes into account the temporal state transitions of the devices inside each window. DTDL shows better precision and recall compared to PED and CDL due to modeling the temporal behavior of consumption signal and learning powerful nonlinear features to boost the disaggregation accuracy.

Fig. 6 shows the actual/estimated power consumption obtained by DTDL for two devices in House 1 on day 14. Notice that the model accurately understands the transients and various steady states in the appliances. Moreover, Fig. 7 depicts the pie charts of the actual/estimated energy consumption of our model and PED, for House 3 during the test time. Notice that our estimated consumption values closely follow the actual values, achieving better accuracy compared to PED in the 7-day test period. This shows better reliability of our proposed model for real-world energy disaggregation purposes in long time horizons.

TABLE II
PRECISION(%), RECALL(%), AND F-SCORE(%) COMPARISONS

| Method | Metric | House 1 | 2 | 3 | 4 | 5 | Average |
|---|---|---|---|---|---|---|---|
| SMP | P | 37.78 | 36.52 | 35.42 | 36.69 | 36.73 | 36.62 |
|  | R | 35.51 | 37.64 | 39.71 | 42.41 | 40.75 | 39.20 |
|  | F-score | 36.60 | 37.07 | 37.44 | 39.34 | 38.63 | 37.82 |
| FHMM | P | 77.12 | 68.81 | 67.63 | 71.83 | 70.09 | 71.09 |
|  | R | 53.45 | 50.02 | 51.54 | 54.59 | 52.88 | 52.49 |
|  | F-score | 63.13 | 57.92 | 58.49 | 62.03 | 60.28 | 60.37 |
| AMAPI | P | 80.56 | 73.57 | 75.82 | 70.27 | 76.79 | 75.40 |
|  | R | 57.83 | 52.81 | 55.23 | 53.29 | 55.63 | 54.95 |
|  | F-score | 67.32 | 61.48 | 63.90 | 60.61 | 64.51 | 63.56 |
| HieFHMM | P | 80.81 | 77.02 | 74.09 | 67.68 | 83.01 | 76.52 |
|  | R | 58.19 | 54.85 | 55.12 | 52.92 | 59.91 | 56.20 |
|  | F-score | 67.65 | 64.07 | 63.21 | 59.39 | 69.59 | 64.78 |
| CDL | P | 78.02 | 71.27 | 73.58 | 77.90 | 89.82 | 78.11 |
|  | R | 56.79 | 53.02 | 52.11 | 55.54 | 56.05 | 54.70 |
|  | F-score | 65.73 | 60.81 | 61.01 | 64.84 | 69.02 | 64.28 |
| PED | P | 86.03 | 78.89 | 74.05 | 77.12 | 90.02 | 81.22 |
|  | R | 62.29 | 56.70 | 51.23 | 55.51 | 68.22 | 58.79 |
|  | F-score | 72.26 | 65.98 | 60.56 | 64.55 | 77.61 | 68.19 |
| DTDL | P | **90.87** | **85.12** | **75.22** | **93.81** | **94.71** | **87.94** |
|  | R | **63.09** | **60.37** | **52.80** | **68.02** | **68.01** | **62.49** |
|  | F-score | **74.47** | **70.63** | **62.04** | **78.85** | **79.17** | **73.03** |

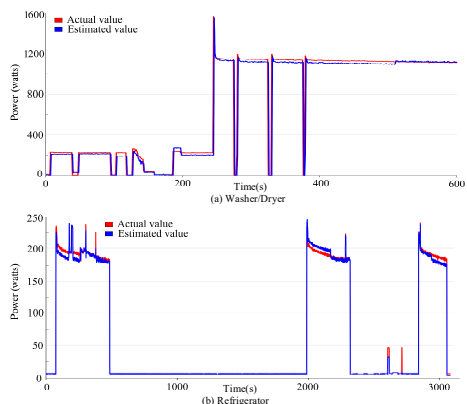

Fig. 6. Estimated energy consumption signals of washer/dryer and refrigerator in House 1 on day 14.

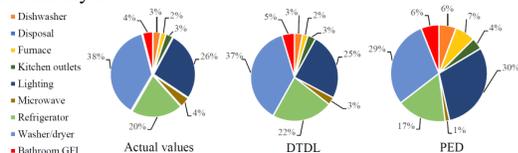

Fig. 7. Pie charts of actual/estimated consumption signals for House 3.

## VI. CONCLUSIONS

In this paper, the problem of energy disaggregation is addressed as a supervised DL problem; A dictionary matrix is learned to capture the representative consumption patterns of each device; Furthermore, a set of coefficients are optimized to find the most accurate sparse linear combination of these patterns to construct the aggregate electricity signal. To extract informative time-dependent electricity patterns, we propose DTDL that learns deep temporal features from the energy signals of each device using an LSTM-AE. An optimization program is devised to learn our LSTM states/parameters while tuning the dictionary atoms and their sparse coefficients using our nonlinear temporal states. Real energy disaggregation experiments on a publicly available dataset show the superiority of our DTDL over HMM-based approaches and dictionary learning models. Compared to the state-of-the-art PED, our DTDL obtains 7.63% and 7.10% better disaggregation accuracy and F-score, respectively. This outperformance is mainly due to extracting nonlinear dictionaries as well as learning temporal structure of the underlying electricity signals. Future research seeks to design a new LSTM-AE whose states can be retrieved by an analytical optimizer such as ADMM-based optimization methods in order to find the global optima temporal parameters.

REFERENCES

[1] A. Rahimpour, H. Qi, D. Fugate and T. Kuruganti, "Non-Intrusive Energy Disaggregation Using Non-Negative Matrix Factorization With Sum-to-k Constraint", *IEEE Transactions on Power Systems,* vol. 32, no. 6, pp. 4430-4441, 2017.
[2] G.W. Hart, "Nonintrusive appliance load monitoring," *Proc. IEEE*, vol. 80, no. 12, pp. 1870–1891, Dec. 1992.
[3] S. Gupta, M. Reynolds and S. Patel, "ElectriSense", *Proceedings of the 12th ACM international conference on Ubiquitous computing - Ubicomp '10*, 2010.
[4] M. Mahmoudi and K. Tomsovic, "A distributed control design methodology for damping critical modes in power systems," in *Proc. Power Energy Conf. Illinois*, 2016, pp. 1–6.
[5] A. Asadinejad andK. Tomsovic, "Optimal use of incentive and price based demand response to reduce costs and price volatility," *Electr. Power Syst. Res.*, vol. 144, pp. 215–223, 2017.
[6] E. Elhamifar and S. Sastry, "Energy disaggregation via learning powerlets and sparse coding," in *Twenty-Ninth AAAI Conference on Artificial Intelligence,* 2015.
[7] M. Mengistu, A. Girmay, C. Camarda, A. Acquaviva and E. Patti, "A Cloud-based On-line Disaggregation Algorithm for Home Appliance Loads", *IEEE Transactions on Smart Grid,* pp. 1-1, 2018.
[8] J. Z. Kolter, S. Batra, and A. Y. Ng, "Energy disaggregation via discriminative sparse coding," in *Proc. 24th Annu. Conf. Neural Inf. Proc. Syst. (NIPS),* Vancouver, BC, Canada, 2010, pp. 1153–1161.
[9] H. Kim, M. Marwah, M. Arlitt, G. Lyon, and J. Han, "Unsupervised disaggregation of low frequency power measurements," in *11th International Conference on Data Mining,* 2010, pp. 747–758.
[10] F. Facchinei, G. Scutari and S. Sagratella, "Parallel Selective Algorithms for Nonconvex Big Data Optimization", *IEEE Transactions on Signal Processing,* vol. 63, no. 7, pp. 1874-1889, 2015.
[11] W. Deng, M.-J. Lai, Z. Peng, and W. Yin, "Parallel multi-block ADMM with O(1/k) convergence," Mar. 2014, arXiv:1312.3040 [Online]. Available: http://arxiv.org/abs/1312.3040.
[12] J. Z. Kolter and M. J. Johnson, "REDD: A Public Data Set for Energy Disaggregation Research," in *Proceedings of the SustKDD Workshop on Data Mining Applications in Sustainability,* 2011.
[13] J. Kolter and T. Jaakkola, "Approximate inference in additive factorial hmms with application to energy disaggregation," *Journal of Machine Learning Research - Proceedings Track,* vol. 22, pp. 1472–1482, 2012.
[14] J. Huang, Z. Zhang, Y. Li, Z. Peng, J. H. Son, "Energy disaggregation via hierarchical factorial hmm," in *Proc. of The Second International Workshop on NILM,* 2014, pp. 1-4.